\documentclass{article}

\usepackage{microtype}
\usepackage{graphicx}
\usepackage{subfigure}
\usepackage{booktabs} % for professional tables

\usepackage{hyperref}
\usepackage{amsmath}

\usepackage[accepted]{icml2020}

\icmltitlerunning{More Information Supervised Probabilistic Deep Face Embedding Learning}

\begin{document}

\twocolumn[
\icmltitle{More Information Supervised Probabilistic Deep Face Embedding Learning}



\icmlsetsymbol{equal}{*}

\begin{icmlauthorlist}
\icmlauthor{Ying Huang}{equal,huya}
\icmlauthor{Shangfeng Qiu}{equal,huya}
\icmlauthor{Wenwei Zhang}{huya}
\icmlauthor{Xianghui Luo}{huya}
\icmlauthor{Jinzhuo Wang}{oxford}
\end{icmlauthorlist}

\icmlaffiliation{huya}{Guangzhou Huya Technology Co., Ltd, Guangzhou, Guangdong, China}
\icmlaffiliation{oxford}{Department of Engineering Science, University of Oxford, Oxford, Oxfordshire, United Kingdom}

\icmlcorrespondingauthor{Ying Huang}{huanying@huya.com}

\icmlkeywords{Face Recognition, Generative Frontal Face}

\vskip 0.3in]



\printAffiliationsAndNotice{\icmlEqualContribution} % otherwise use the standard text.
\begin{abstract}
Researches using margin based comparison loss demonstrate the effectiveness of penalizing the distance between face feature and their corresponding class centers.
Despite their popularity and excellent performance, they do not explicitly encourage the generic embedding learning for an open set recognition problem.
In this paper, we analyse margin based softmax loss in probability view.
With this perspective, we propose two general principles: 1) monotonically decreasing and 2) margin probability penalty, for designing new margin loss functions.
Unlike methods optimized with single comparison metric, we provide a new perspective to treat open set face recognition as a problem of information transmission. 
And the generalization capability for face embedding is gained with more clean information.
An auto-encoder architecture called Linear-Auto-TS-Encoder(LATSE) is proposed to corroborate this finding.
Extensive experiments on several benchmarks demonstrate that LATSE help face embedding to gain more generalization capability and it boost the single model performance with open training dataset to more than $99\%$ on MegaFace test.
\end{abstract}

\section{Introduction}
\label{intro}
Face recognition performance has gained dramatic improvement in recent years. 
Margin based loss functions\cite{schroff2015facenet, liu2017sphereface} play an important role in this process. 
The most common solutions treat face recognition as a classification problem. 
These works utilize deep convolutional network, such as VGGNet \cite{simonyan2014deep} or ResNet \cite{he2016resnet}, to transfer the landmark aligned face images to their corresponding class label.
And comparison loss function\cite{hadsell2006dimensionality} is employed in the learning process, such as softmax cross entropy loss.
Recent researches demonstrate that we can gain more discriminative face embedding by explicitly adding margin to loss.
FaceNet \cite{schroff2015facenet} adds margin by penalizing the distance between face feature and their corresponding class centers in euclidean space (triplet loss). 
While SphereFace \cite{liu2017sphereface} penalizes this distance in hyperspherical angle space (angular margin softmax loss).

\begin{figure}[t]
%\vskip 0.2in
\begin{center}
\includegraphics[width=\columnwidth]{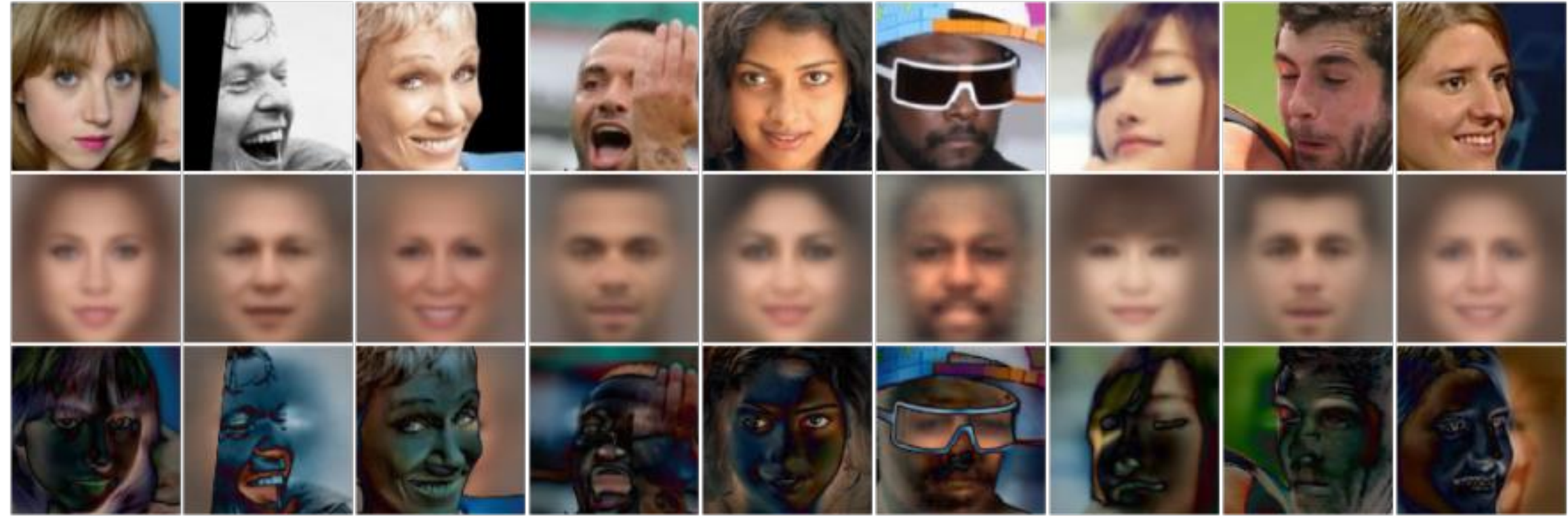} % Reduce the figure size so that it is slightly narrower than the column. Don't use precise values for figure width.This setup will avoid overfull boxes. 
\caption{Examples of generated frontal face from different identity. In rows top to bottom: original input face image, the generated frontal face image from the corresponding feature, the absolute difference between input image and the generated one.}
\label{fig_genface}
\end{center}
%\setlength{\abovecaptionskip}{-0.8cm} 
%\vskip -0.2in
\end{figure}

However, all these methods strengthen the embedding learning process from the perspective of comparison.
So these recognition algorithms optimize single discriminative ability metric. 
At the same time, they do not give a general guidance for new margin loss function designing.

While robust face embedding learning is an open set problem\cite{geng2018recent,boult2019learning}. 
We can not involve images of all possible human identities in the training dataset. 
So we seek to solve some primary questions: \textit{How to gain more discriminative power with appropriate margin definition? Whether discriminative ability equals generative capability in open set situation?}

Following these questions, we firstly explore the effect of margin in softmax cross entropy loss \cite{liu2016large,wang2018cosface,deng2019arcface} from probabilistic view.
Through detailed analysis, this work proposes two principles to guide the definition of new margin loss function. 
One is monotonically decreasing. 
Let $\theta$ be the angle between face feature and their class center. 
The definition of transformation function, which transforms $\theta$ to probability value, should monotonically decrease in the domain of $\theta$.
The other is margin probability penalty.
New margin loss should guarantee a non-negative probability penalty following this principle.

And we find single comparison metric may not direct the model to gain optimally generalized face embedding. 
In contrast to these single-metric-based methods, we investigate a different way. 
In our work, we treat face recognition as an information transmission problem, which can be supervised by class label. 
Different from the classification task which only strengthens the discriminative power, information transmission can be seen as a multi-objective optimization problem.
One optimization objective is comparison-metric-based discriminative embedding learning. 
While the other objective is that how much accurate information is passed to the face feature.

Base on this motivation, we built the learning framework as an auto-encoder \cite{hinton2006reducing} architecture, which can be trained with a teacher student \cite{hinton2015distilling,liu2017iterative} learning strategy.
This Linear-Auto-TS-Encoder(LATSE) architecture can guarantee the proper information pass through the whole network. 
Different parts in framework promise distinct effects for the embedding learning.
In the decode part is a generative network \cite{goodfellow2014generative,zhao2016energy} which transfers face embedding to the corresponding individual frontal face image.
This generative model is supervised by pixel level image label. 
The generated frontal faces are shown in Figure \ref{fig_genface}.
At the encode part, we employ a deep ResNet to learn face embedding. 
A new definition of margin loss following the proposed principle plays the role as comparison loss.
The goal of teacher network is filtering noise and ensuring cleaner information for face embedding learning. 
The effects of these parts are discussed in details below.
We conduct extensive experiments on large scale face recognition benchmarks.
The experimental results verify our findings and the effectiveness of the proposed architecture.

\section{Related Work}
\label{Related}
\textbf{Margin Based Comparison Loss.} 
Learning face embedding with comparison loss can be divided into two main streams. 
One stream methods directly obtain face embedding from the raw image through comparing match/non-match pairs, such as triplet loss \cite{schroff2015facenet}. 
The other methods first train a multi-class CNN using margin-based softmax loss function, such as large-margin softmax loss \cite{liu2016large} or arccos loss \cite{deng2019arcface}. 
Then the feature layer before softmax is used as face embedding. 
Recent works mainly focus on how to adjust margin or other hyper parameters. 
Liu \cite{liu2019adaptiveface} thinks that the margin between unbalance face classes should be adaptively learned during the training process. 
RegularFace \cite{zhao2019regularface} is proposed which explicitly penalize the angle between an identity and its nearest neighbor in order to increase the inter-class separability. 
AdaCos \cite{zhang2019adacos} is proposed to tune the margin and scale parameter automatically to strengthen the discriminative power for face embedding.
However, the single optimization metric, which enhances the intra-class compactness and inter-class dispersion, in all these methods may ignore the nature of open set recognition problem. 

\begin{figure*}[tb]
\centering
\includegraphics[width=0.9\textwidth]{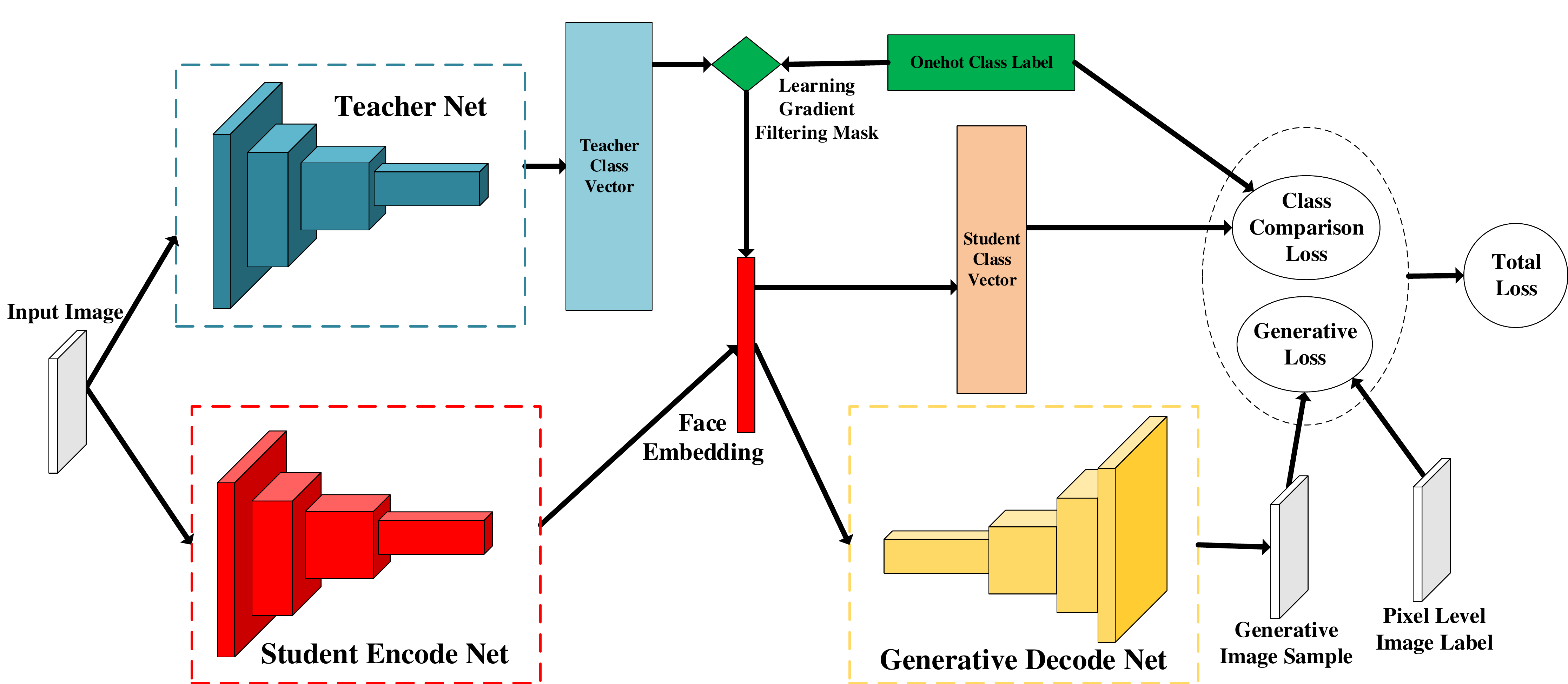} % Reduce the figure size so that it is slightly narrower than the column. Don't use precise values for figure width.This setup will avoid overfull boxes. 
\caption{The proposed Linear-Auto-TS-Encoder(LATSE) Architecture. The red bottom part is the student encode net. While the yellow part is generative decode net. The top blue part is teacher network. A linear comparison loss and a pixel level generative loss are employed for final face embedding learning.}
\label{fig_arch}
\end{figure*}

\textbf{Generative Model.} 
Generative network learning usually starts with a latent code from noise distribution. 
Then the model maps this code to a generative sample, such as probabilistic GAN \cite{goodfellow2014generative} and Energy Based GAN \cite{zhao2016energy}. 
In Cycle GAN \cite{zhu2017unpaired}, a cycle-consistent loss is proposed to learn from unpaired images. 
However, few work explores whether a generative model can enhance the embedding generalization ability in the open set recognition situation. 

\textbf{Teacher Student (TS) Learning Strategy.}
Noises are inevitable in large-scale datasets and heavily affect the performance of face recognition algorithms\cite{wang2018devil}.
It is expensive to get extremely clean large-scale datasets.
So how to learn with noises plays a significant role for model training.
MentorNet \cite{jiang2017mentornet} utilizes a pre-trained teacher network to drop and update corrupted labels for the student network in learning process.
De-Coupling \cite{malach2017decoupling} alleviates influence of  noises through updating the parameters only when the predictions from two classfiers are same.
Co-Mining \cite{wang2019co} simultaneously traines two peer networks to redistribute the labels of raw data.
Despite these methods have made efforts to reduce the corruption of noisy labels. 
There are still drawbacks to be improved, such as careful design for face recognition with large number of classes, less resource consumption and easier training process.

\textbf{Multi-Modal Supervision For CNN Models.} 
CNN for Image recognition is generally trained with image level semantic labels \cite{krizhevsky2012imagenet}. 
While several researches demonstrates that multi-modal supervision can strengthen the generative ability of the learned models in different types of tasks. 
MTCNN \cite{zhang2016joint} and RetinaFace \cite{deng2019retinaface} find that face detection performance can be boosted by using face landmarks along with bounding boxes label. 
Mask-rcnn \cite{he2017mask} boosts the performance for the task of instance segmentation, bounding-box object detection, and person keypoint detection by utilizing object level bounding boxes and pixel level semantic labels together. 
Depth and time related information is used in Depth-Patch-Net \cite{atoum2017face} and AuxNet \cite{liu2018Aux} to improve the accuracy for the task of anti-face spoofing. 
However, there is few work to explore whether face recognition performance can be improved by extra supervision. 
In this work, we leverage pixel level face image as extra supervision to improve face recognition accuracy.

\section{Learning Face Feature Embedding with Multi-Modal Supervision}
\label{Learning}
The intuition behind this work is more orthogonal information lead to less uncertainty.  
And we treat face recognition as a problem of information transmission. 
Following this guide, we build our learning architecture as an  auto-encoder framework which is trained with a teacher student learning strategy. 
This architecture is illustrated in Figure \ref{fig_arch}.
Although there are three key components in the proposed algorithm. 
The model can be trained end-to-end from scratch data.
In the following paragraphs,
we will first illustrate the probability theory behind margin comparison loss. 
Then details for multi-level supervision are gave. 
After that is teacher student learning strategy and whole network training algorithm.

\subsection{Margin Softmax Loss in Probability View}
\label{LearningMargin}
There is a line of research to add margin in softmax cross entropy loss. 
Previous works illustrate their theory by giving us a geometric interpretation \cite{liu2016large,deng2019arcface}.
In this paper, we try to explain the effect of margin in probability view. 
At the same time, we summarize two principles which can be followed when new margin loss is needed.  

Firstly, we starts from the most widely used softmax loss function. 
Let $x_i$ represent the extracted feature for a face image from identity $y_t$. 
$W_{y_t}$ and $W_j$ are the model weights in the fully-connected classifier layer for identity $y_t$ and $y_j$. 
$b_{y_t}$, $b_j$ are the biases. 
$N$ is the total image number and $K$ is the total category number in the training set.
Then the softmax loss function can be presented as:
\begin{equation}
    L = -\frac{1}{N} \sum_{i=1}^N\log\frac{e^{{W_{y_t}^Tx_i+b_{y_t}}}}{\sum_{j=1}^Ke^{{W_j^Tx_i+b_j}}}
    \label{eq_softmax_ce}
\end{equation}
This loss is consisted of two components and each part plays a distinct role in model optimization.
One is the softmax function which transforms the predicted value from fully-connected layer to the probability of corresponding class.
The other is a cross entropy loss to measure the difference between predicted probability and the given label distribution. 
Then we can decouple the softmax loss in Equation \ref{eq_softmax_ce} to these parts.
For an input image $x$, the softmax function computing the probability for the target label $y_t$ can be presented as:
\begin{equation}
    P_{softmax}(x_i, y_t) = \frac{e^{{W_{y_t}^Tx_i+b_{y_t}}}}{\sum_{j=1}^Ke^{{W_j^Tx_i+b_j}}}
\end{equation}
Let $q(x)$ represent the real data distribution and $p(x)$ be the model predicted probability distribution. 
Then the cross entropy loss computes the mutual information $H(p,q)$ between them, which can be presented as:
\begin{equation}
    CELoss = H(p,q) = -\sum_x q(x)\log p(x) 
\end{equation}
Following the researches for margin softmax loss \cite{liu2017sphereface,deng2019arcface}, we do not modify the format of cross entropy loss. 
And we adjust margin in softmax function. 
For simplicity, we fix the bias term $b_j=0$ as in \cite{liu2016large}, normalize the feature $\lVert x_i \rVert = 1$ and classifier layer weight $\lVert W_j \rVert = 1$. 
Then the predicted value from fully-connected layer $W_j^Tx_i+b_j$ is simplified as $\lVert x_i \rVert \lVert W_j \rVert \cos{\theta}$.
After that, the output is multiplied with a scale parameter $s$. 
At this time, softmax function is formulated as:
\begin{equation}
    P_\theta(x_i, y_t) = \frac{e^{s(\lVert x_i \rVert \lVert W_{y_t} \rVert \cos\theta_{y_t})}}{\sum_{j=1}^Ke^{s(\lVert x_i \rVert \lVert W_j \rVert\cos\theta_j)}} = \frac{e^{s\cos\theta_{y_t}}}{\sum_{j=1}^Ke^{s\cos\theta_j}}
\end{equation}
where $\theta_j$ is the angle between the predicted face feature and its normalized class center. 
In previous works \cite{liu2018learning,chen2019angular}, the angle $\theta$ is demonstrated to be correlated to visual semantic.
Decreasing this angle can boost the discriminative ability for the learned model.
Margin based softmax methods try to multiple ($m_1$) or add positive margin value ($m_2$ or $m_3$) in the target label, which is expressed in format:
\begin{equation}
    P_m(x_i, y_t) = \frac{e^{s(\cos(m_1\theta_{y_t}+m_2)-m_3)}}{e^{s(\cos(m_1\theta_{y_t}+m_2)-m_3)}+\sum_{j\ne{y_t}}^Ke^{s\cos\theta_j}}
\end{equation}
Rethinking the effect of softmax function, it normalizes the predicted value $W_j^Tx_i+b_j$ from the fully-connected layer and transfers this value to the probability of the corresponding class. 
Then we can define a general function $F$ to normalize the predicted value:
\begin{equation}
    F(x_i, y_t) = \frac{\phi(x_i, W_{y_t})}{\sum_{j=1}^K\phi(x_i, W_j)}
\end{equation}
Where $\phi(x, W) >=0 $ should always hold for any input pair $(x, W)$. 
At the same time, the definition of this generalized function $F$ must satisfy probability law. 
While in softmax function, it employs $\phi(x,W) = e^{W^Tx}$ as the definition of $\phi$.
By adding margin $m$ in this normalization function $F$,  we can get $F_m$ :
\begin{equation}
    F_m(x_i, y_t) = \frac{\chi(x_i, W_{y_t}, m)}{\chi(x_i, W_{y_t}, m)+\sum_{j\ne{y_t}}^K\phi(x_i, W_j)}
\end{equation}
Where $\chi$ and $\phi$ are general functions to transform fully-connected layer output to a non-negative value.

\textbf{The first principle for the definition of $\chi$ and $\phi$ is non-negative and to penalize probability when margin is added.} 
This principle is formulated as:
\begin{equation}
\phi(x,W) \ge \chi(x, W, m) \ge 0
\end{equation}
When we have non-negative margin value $m_1$ and $m_2$,  $\chi$ should be monotonically decreasing when the margin increases. 
\begin{equation}
\chi(x, W, m_1) \le \chi(x, W, m_2),   m_1 \ge m_2 \ge 0
\end{equation}
Margin based loss decreases the probability of target label when the model predicts same angle between feature and class center, expressed as $F_m \le F$. 

\textbf{Another principle for function $\chi$ and $\phi$ is to make them monotonic decrease in the domain of $\theta$} ($x_iW_j^T=\cos\theta_j$ when $\lVert x_i \rVert = \lVert W_j \rVert = 1$). 
If the model tries to get the same probability under a margin based function, it must step forward to decrease the angle between feature and class center.
This constraint makes the learned model gain more discriminative power by learning small angle $\theta$ between similar inputs.

Following these proposed principles, we can generalize the margin based probability function $F_m$ to any new formats.
In this paper, we test our method with a linear definition under these principles, specialize $\chi = e^{h(\theta)}$ and $h(\theta)= s(-a\theta + b)$.  We set $\phi = e^{g(\theta)}$ and $g(\theta) = {s\cos\theta}$.
This specialization case can be formulated as:
\begin{equation}
P_{linear}(x_i, y_t) = \frac{ e^{s(-a\theta_{y_t} + b)}}{e^{s(-a\theta_{y_t} + b)}+\sum_{j\ne{y_t}}^Ke^{s\cos\theta_j}}
\end{equation}
Finally, the $DLoss$ is computed by cross entropy loss function between model prediction $P_{linear}$ and label probability $q$. This is defined as:
\begin{equation}
DLoss(x) =  -\sum_x q(x)\log P_{linear}(x)
\end{equation}

\begin{figure}[ht]
%\vskip 0.2in
\begin{center}
\includegraphics[width=0.9\columnwidth]{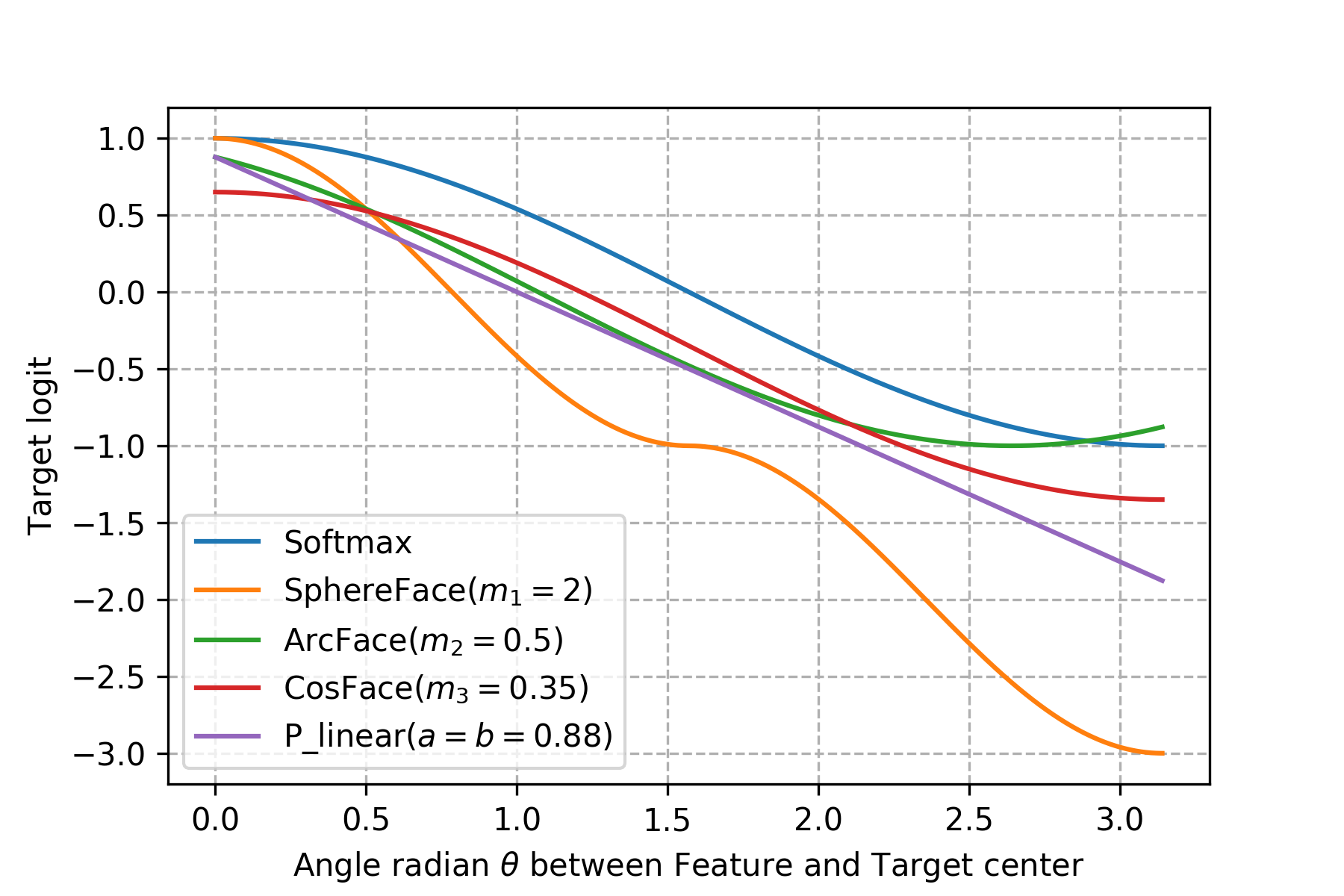} % Reduce the figure size so that it is slightly narrower than the column. Don't use precise values for figure width.This setup will avoid overfull boxes. 
\caption{Target logit curves for different loss function}
\label{fig_target}
\end{center}
\vskip -0.2in
\end{figure}

\textbf{Visualize the effect of margin by target logit curve.} 
Different kinds of margin try to decrease the predicted probability for ground truth by penalizing the target predicted value, namely increase the cross entropy loss value under the same angle $\theta$. 
Compared to the origin softmax function, we can view their relations in Figure \ref{fig_target}.
Despite the similar goal for different margin losses, the proposed principles will lead us to define a better normalization function.  
They provide a guidance when problem occurs in model training. 
From target logit curve in the domain of $\theta$, we can find SphereFace\cite{liu2017sphereface} has a nonlinear penalty value which increases along with angle $\theta$. 
This leads divergence at the beginning of training when the initial angle $\theta$ is large. 
Arcface\cite{deng2019arcface} is not monotonic decreasing in the domain of $\theta$, so it will shrink margin instead of magnifying it when angle is too large. 
CosFace\cite{wang2018cosface} is too smooth in the domain of $\theta \in (0, 0.5)$, meaning it will have little loss decay if the angle $\theta$ becomes smaller in this area. 
This makes it hard to shrink angle at the final stage of training.
By contrast, the proposed linear format stably increases the target predicted value along with angle decreasing. 
This setting reduces the difficulty for the model to learn small angle $\theta$ between face embedding and its class center.

\subsection{Pixel Level Supervision for Face Generative Net}
\label{LearningGan}
Owing to single classification optimization metric (intra-class compactness and inter-class dispersion), margin based comparison loss encounters the bottleneck to gain more generalization ability for face embedding learning. 
We propose whether model can transfer more information to the face feature as another metric, which also should be optimized in the embedding learning process.  
So a generative network is added to realize this purpose with pixel level supervision. 
This generative part will try to restore the frontal face image from the embedding. 
Therefore the learned face embedding need to keep as much information as possible. 
Although we train this generative part with pixel level supervision. 
It is gained without cost of extra human labeling expect the identity class label for the whole image. 
The loss to supervise the generative model is consisted of two components. 
One is $L1Loss$ for frontal face regression.
The other is $SimLoss$ correlated to structural similarity index \cite{wang2004image}, which compares the similarity between generative sample and the input image. 
Given an face image $x^c$ for identity $c$. 
The generated face is $x^c_{gen}$. 
And the label for generative model $y^c_{img}$ is the momentum mean image from $c$, which can be expressed as:
\begin{equation}
    y^c_{img} = (1-momentum) * x^c + momentum * y^c_{img}
\end{equation}
Then We gain the definition of $SimLoss$:
\begin{equation}
    SimLoss(x^c_{gen}, y^c_{img}) = \frac{1 - SSIM(x^c_{gen},  y^c_{img})}{2}
\end{equation}
Next the total generative loss $GLoss$ is expressed as:
\begin{equation}
    GLoss(x^c_{gen}, y^c_{img}) = L1Loss + SimLoss
\end{equation}
The $L1loss$ suggest the model to transfer more information to embedding. 
While $SimLoss$ make the generated image to resemble the frontal face of the input identity. 

\subsection{Teacher Student Learning Strategy}
\label{LearningTeacher}
Nowadays, most of large-scale datasets are obtained by utilizing search engine. 
Researchers apply automatic or semi-automatic approaches to clean the identity label for these datasets.
This process leads to two types of noise : 1) label mess, which means images from same person may be marked with different labels. 2) distractor, which means some labeled examples are not part of any class within the dataset.
Both types of noise will pollute the generalization ability of the learned face embedding.
Despite current methods \cite{jiang2017mentornet, malach2017decoupling, wang2019co} have provided various strategies for model training in noisy dataset.
Some of them are designed for binary classification and others may make training process too complicated.
To solve these deficiencies and take into account the character of face embedding, we propose a teacher student learning strategy to filter label noise real-time in training process.

This strategy uses only the correct predictions gained by the pre-trained teacher model to update the student network, which guarantees training samples sufficiently clean with the prior knowledge of the teacher network. 
Recent researches \cite{yu2019does, NIPS2018_8072} mention the memorization effects of deep neural networks, which argue that networks would first memorize training data of clean labels.
So we train a teacher network in the large-scale dataset to make the teacher network memorize data of clean labels as much as possible and then fix its parameters in next process.
Then, the teacher network will filter noise for the student network, which would further strengthen representational ability of the face embedding.
The filtering strategy of the teacher network is optional, which provides more flexibility.
Besides, there is no need to simultaneously train the teacher and  student model.
This setting simplifies the procedure of training and achieves better results in consumption of training time and resource than other methods.

Intuitively, the more correct knowledge gained from teacher, student can form more distinct cognition about the problem.
Equally like that in the training process, the face embedding learned by the student network would possess better discriminability and generalization with the clean data under the help of the auxiliary teacher network. 
Besides, in the view of information transmission, the teacher net plays role as high reliability signal channel between face image data distribution and identity label.

\subsection{Whole network training}
For the whole network training. 
First we gain a teacher model on training data. 
Then the parameters from teacher network are fixed. 
Next we estimate the student network and generative model on same dataset with $Loss$ formulated as:
\begin{equation}
Loss(x) =  DLoss(x)+GLoss(x)
\end{equation}
Finally the parameters from student model are saved for testing. 
The whole training process can be illustated by Algorithm\ref{alg:latse}.
\begin{algorithm}[tb]
   \caption{Linear-Auto-TS-Encoder Learning algorithm}
   \label{alg:latse}
\begin{algorithmic}
   \STATE {\bfseries Input:} face images $X$, face class label $Y$
   \STATE First train $teacherNet$ by
   \FOR{$i=1$ {\bfseries to} $maxiter$}
     \STATE Learn $teacherNet$ with $X$ and $Y$ using $DLoss$
   \ENDFOR
   \STATE Then fix parameters of $teacherNet$
   \STATE Next Estimate $studentNet$, $genNet$ with $X$ and $Y$ by
   \FOR{$i=1$ {\bfseries to} $maxiter$}
     \STATE $prob_{yteacher} = teacherNet.forward(x_i)$ 
     \STATE $em_{student} = studentNet.forward(x_i)$
     \STATE $prob_{student}$= $FCLMSoftmax.forward(em_{student})$
     \STATE $DLoss = CELoss(prob_{student}, y_i)$
     \STATE $xgen_i = genNet.forward(em_{student})$ 
     \STATE $GLoss = GLossFun(xgen_i, x_i)$
     \STATE $Loss = DLoss + GLoss$
     \STATE $gradGLoss = genNet.backward(GLoss)$
     \STATE $gradDLoss = FCLMSoftmax.backward(DLoss)$
     \STATE $gradFromLoss = gradDLoss+gradGLoss$
     \IF{$y_i \in topk(prob_{yteacher},k)$}
     \STATE $gradForEm$ = $filter(gradFromLoss)$
     \ENDIF
     \STATE $studentNet.backward(gradForEm)$
   \ENDFOR
   \STATE Finally save $studentNet$ parameters for testing
\end{algorithmic}
\end{algorithm}

\section{Experiments}
\subsection{Implementation Details}
\textbf{Datasets.} We employed CASIA \cite{yi2014learning} as small training set and MS1MV2 \cite{guo2016ms} or MS1M-RetinaV \cite{deng2019arcface} from ArcFace \cite{deng2019arcface} as large training dataset. 
We compared the performance for both face verification and identification tasks on several benchmark datasets, including Labelled Faces in the Wild (LFW) \cite{huang2008labeled}, Celebrities in Frontal Profile  (CFP-FP) \cite{sengupta2016frontal}, Age Database (AgeDB-30) \cite{moschoglou2017agedb}, Cross-Age LFW (CALFW) \cite{zheng2017cross}, Cross-Pose LFW (CPLFW) \cite{zheng2018cross} and Megaface \cite{nech2017level}. 

\textbf{Experimental Settings.}
The proposed method were implemented with MXNet \cite{chen2015mxnet}.
We reduced memory cost with the help of memonger \cite{chen2016training}. 
The data preprocessing step followed paper of margin based softmax loss \cite{liu2017sphereface, wang2018cosface, deng2019arcface}. 
The detected face were aligned by five facial key points and resized to fix dimension ($112 \times 112 \times 3$) as the network input. 

ResNet \cite{he2016resnet} with depth of 34, 50, 100 and 124 were employed as the encode part in the auto-encoder, followed by a structure of Batch Normalization \cite{ioffe2015batch}, Dropout, Fully-Connected layer and Batch Normalization to get the final 512 dimension face embedding. 
A reversed ResNet-18 with deconvolution layer \cite{noh2015learning} to up-sample feature map was adopted as the generative decode part. 
A same setting with the encode part was employed for teacher network. 
The parameters in teacher network were fixed during the student network training process. 

For hyperparameter setting, we adopted $s=64$ as the normalized scale in spherical manifold by following \cite{liu2017sphereface}. 
Models were trained in eight NVIDIA Tesla V100 GPUs(16GB) with total batch size 768. 
The learning rate started from 0.1 and was divide by 10 at 10K, 16K, 20K, 22k iterations. 
We set weight decay to $5e^{-4}$ and momentum to 0.9. 
At test time, we only computed the 512 dimension feature for each normalized face from the student network and compared feature cosine angle value as the similarity score between different face images. 

\begin{table}[ht]
\caption{Face verification results ($\%$) with different margin loss.(models trained on CASIA, ResNet50)}
\label{loss_table}
%\vskip 0.15in
\begin{center}
\begin{small}
%\begin{sc}
\begin{tabular}{lcccr}
\toprule
Loss Functions & LFW & CFP-FP & AgedDB-30 \\
\midrule
\scriptsize{ArcFace \cite{deng2019arcface}} & 99.53 & 95.56 & $\mathbf{95.15}$ \\
\scriptsize{SphereFace \cite{liu2017sphereface}} & 99.42 & - & - \\
\scriptsize{CosFace \cite{wang2018cosface}} & 99.51 & 95.44 & 94.56 \\
\midrule
\scriptsize{LinearFace$(a=0.88,b=0.88)$}   & $\mathbf{99.55}$ & $\mathbf{97.01}$ & 94.66 \\
\scriptsize{LinearFace$(a=0.88, b=1)$}     & 99.48 & 96.81 & 95.05 \\
\scriptsize{LinearFace$(a=1, b=1)$}     & 99.48 & 96.80 & 94.40 \\
\bottomrule
\end{tabular}
%\end{sc}
\end{small}
\end{center}
%\vskip -0.1in
\end{table}

\subsection{Ablation Study of proposed key components}
\textbf{The performance with different margin loss.} 
Firstly, we explored the effect of different margin losses. 
For fair comparison, the proposed LinearFace method was trained following the setting in \cite{deng2019arcface} with ResNet50 as embedding feature backbone on CASIA dataset. 
The verification results on LFW, CFP-FP and AgeDB-30 were reported to compare their performance in Table \ref{loss_table}. 
From the result, we observed better accuracy on LFW and CFP-FP by employing the linear margin function compared to other margin softmax losses, which demonstrated the effectiveness of the proposed principle in Section \ref{LearningMargin}.

\begin{table}[ht]
\caption{Face verification results ($\%$) with Teacher Student learning strategy}
\label{ts_table_one}
\vskip 0.15in
\begin{center}
\begin{small}
%\begin{sc}
\begin{tabular}{lcccr}
\toprule
Method & LFW & CALFW & CFP-FP \\
\midrule
\scriptsize{resnet34} &99.65  &95.85  &92.12  \\
\scriptsize{resnet34(k=1)} & $\mathbf{99.76}$  &96.03  &97.23  \\
\scriptsize{resnet34(k=3)} & $\mathbf{99.76}$  & $\mathbf{96.05}$  & $\mathbf{97.27}$  \\
\midrule
\scriptsize{resnet50} &99.80  &95.80  &92.74  \\
\scriptsize{resnet50(k=1)} & $\mathbf{99.81}$  &95.93  &98.25  \\
\scriptsize{resnet50(k=3)} & $\mathbf{99.81}$  & $\mathbf{95.95}$  & $\mathbf{98.30}$  \\
\midrule
\scriptsize{resnet100} &99.77  &$\mathbf{96.10}$  &98.27  \\
\scriptsize{resnet100(k=1)} & $\mathbf{99.83}$  & $\mathbf{96.10}$  &98.64  \\
\scriptsize{resnet100(k=3)} & $\mathbf{99.83}$  & $\mathbf{96.10}$  & $\mathbf{98.71}$  \\
\bottomrule
\end{tabular}
%\end{sc}
\end{small}
\end{center}
\vskip -0.1in
\end{table}

\begin{table}[ht]
\caption{Megaface performance ($\%$) with Teacher Student learning strategy}
\label{ts_table_two}
\vskip 0.15in
\begin{center}
\begin{small}
%\begin{sc}
\begin{tabular}{lcccr}
\toprule
Method          & Identification  & Verification \\
\midrule
\scriptsize{resnet34} &96.09  &96.72   \\
\scriptsize{resnet34(k=1)} &97.61  &98.03   \\
\scriptsize{resnet34(k=3)} & $\mathbf{97.74}$  & $\mathbf{98.03}$   \\
\midrule
\scriptsize{resnet50} &97.26  &97.62    \\
\scriptsize{resnet50(k=1)} &98.26  &98.48    \\
\scriptsize{resnet50(k=3)} & $\mathbf{98.33}$  & $\mathbf{98.55}$   \\
\midrule
\scriptsize{resnet100} &98.35  &98.6   \\
\scriptsize{resnet100(k=1)} &98.56  &98.8   \\
\scriptsize{resnet100(k=3)} & $\mathbf{98.70}$  & $\mathbf{98.83}$   \\
\bottomrule
\end{tabular}
%\end{sc}
\end{small}
\end{center}
\vskip -0.1in
\end{table}

\begin{figure}[ht]
%\vskip 0.2in
\begin{center}
\includegraphics[width=0.9\columnwidth]{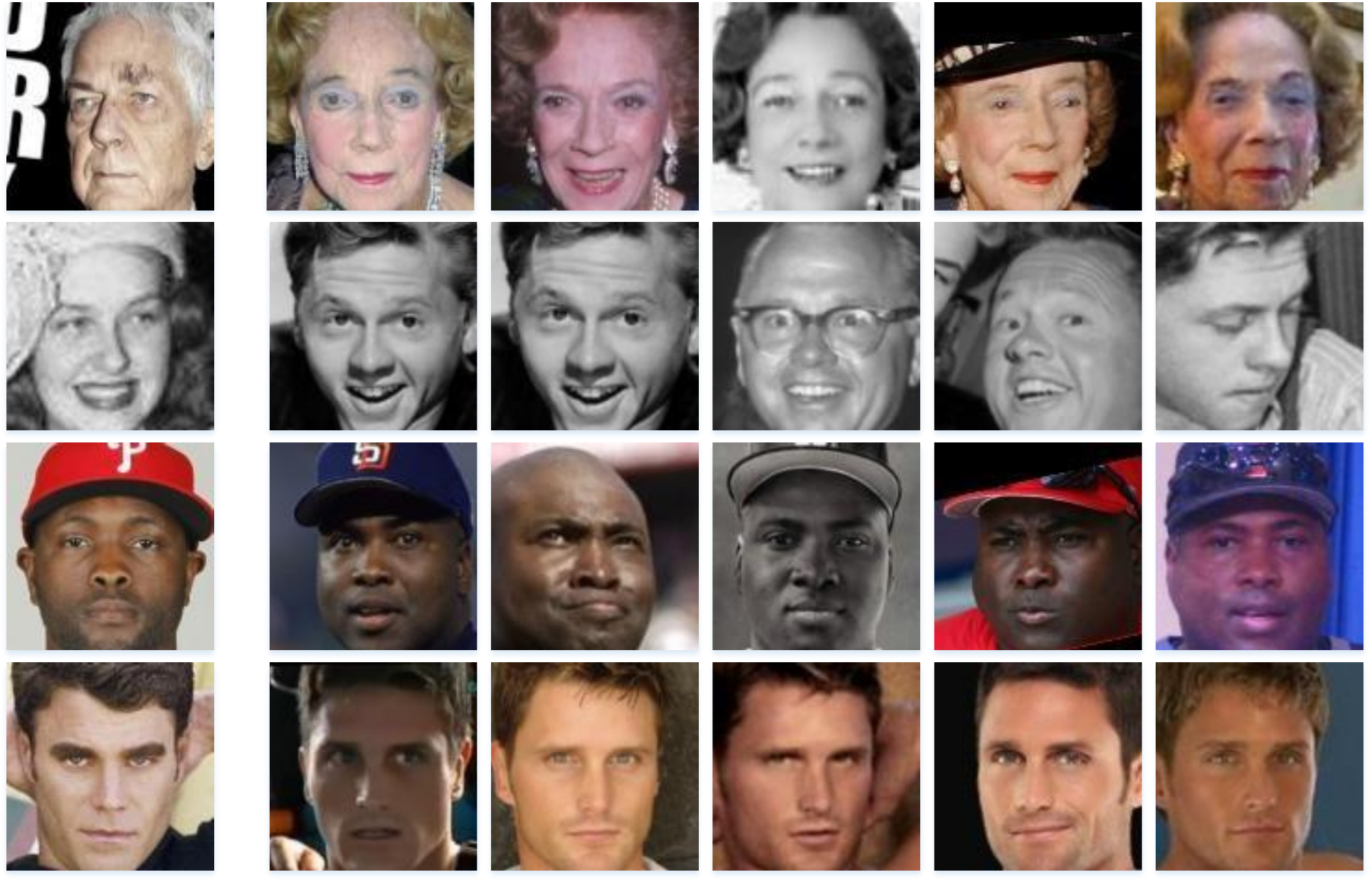} % Reduce the figure size so that it is slightly narrower than the column. Don't use precise values for figure width.This setup will avoid overfull boxes. 
\caption{Samples selected by teacher network. Examples from same row were marked with same label. The leftmost column: data of true label. The other columns: data of noisy labels. The student network actually avoid pollution of noisy labels with the selection of teacher network.}
\label{fig_badcases}
\end{center}
%\vskip -0.2in
\end{figure}

\textbf{The effectiveness of Teacher Student Strategy.} 
To show the effectiveness of teacher student learning strategy, we conducted numerous validity experiments. 
All experiments employed MS1MV2 as training dataset and Arcface loss as training loss to guarantee fairness of comparison.
We set the model performance from \cite{deng2019arcface} as baseline.
The performance for student network was explored with setting different $k$ values of teacher network.
We assume the image label was correct if it was included in teacher network predicted top $k$ probabilities.
Only when image label fell in teacher's predicted top $k$ classes, the gradient would pass to student network for updating its parameters. 
As shown in Table \ref{ts_table_one} , our teacher student learning strategy could be applied effectively with various network depth(34,50 and 100).
The strategy not only obtained better results in several benchmarks, especially got more than $6\%$ promotion in CFP-FP when depth was 50. 
But also outperformed baseline for large-scale evaluation set. Results in Table \ref{ts_table_two} verified that our strategy strengthen the generalization ability of face embedding.
Besides, we made visualization of clean training samples provided by teacher network during student training process in Figure \ref{fig_badcases}.
Obviously, the student network could get cleaner samples in large-scale training dataset with the help of the teacher network.

\begin{table}[t]
\caption{The effect with different proposed component}
\label{component_table}
\vskip 0.15in
\begin{center}
\begin{small}
%\begin{sc}
\begin{tabular}{lcccr}
\toprule
Component & CALFW & CPLFW  & AgedDB-30 \\
\midrule
\scriptsize{LinearFace} & 93.3 & 89.08 &  94.66 \\
\scriptsize{LinearFace+TS} & 93.73 & 89.75 & 95.05  \\
\scriptsize{LinearFace+Gen} & 94.3 & 90.13 & $\mathbf{95.91}$ \\
\scriptsize{LinearFace+TS+Gen} & $\mathbf{94.58}$ & $\mathbf{90.33}$  & $\mathbf{95.91}$ \\
\bottomrule
\end{tabular}
%\end{sc}
\end{small}
\end{center}
\vskip -0.1in
\end{table}
\textbf{The advantage of proposed architecture.} 
Next we compared four architectures with different components equipped in them to show the advantages of the proposed architecture. 
The face verification performance on different benchmark datasets is shown in Table \ref{component_table}. 
LinearFace represents an architecture equipped with only the proposed margin loss. 
TS is short for the teacher student learning strategy and Gen is short for the generative decode part. 
The row LinearFace+TS+Gen showed the performance of the proposed architecture in this paper.  
All these model were trained on CASIA dataset with ResNet50 as backbone.
The results showed that the proposed architecture can effectively boost the performance for face verification.

\begin{figure}[ht]
%\vskip 0.2in
\begin{center}
\includegraphics[width=\columnwidth]{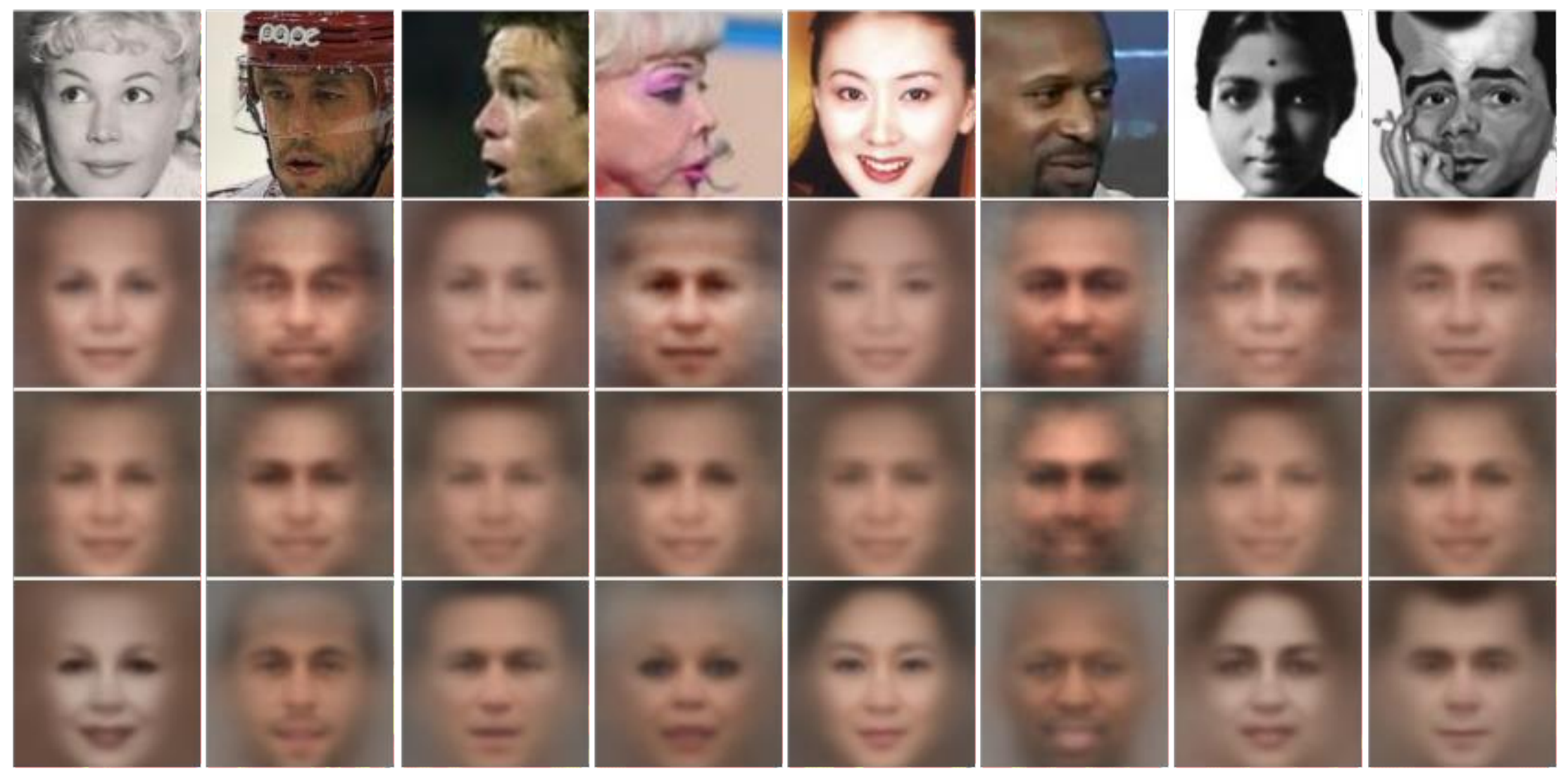} % Reduce the figure size so that it is slightly narrower than the column. Don't use precise values for figure width.This setup will avoid overfull boxes. 
\caption{Examples of generated frontal face for same identity from different embedding models of various training stage. In rows top to bottom: early stage embedding model, middle stage embedding model and last stage embedding model. More details reserved in the generated frontal face images from latter stage embedding model.}
\label{fig_stage}
\end{center}
%\vskip -0.2in
\end{figure}

\textbf{Visualize samples from generative model.}
Resent researches illustrated why margin based algorithm works for face embedding learning by giving an geometric interpretation.
But the perceptual intuition what the feature learned through the embedding network was missing.
With the help of the auto-encoder architecture, we have a chance to visualize what different embeddings look like directly. 
In Figure \ref{fig_stage}, we showed the generative samples for same identity. 
We employed different embedding models from early, middle and last stages of training as the encode model to get the face image feature.
Then a pre-trained generative decode net was employed to restore these samples from features.
The figure showed that along with the training process stepping forward, more details were preserved in the restore frontal face image. 
This suggested that the embedding model has gained more discriminative power. 
And the model increased the capacity to retain information.

\subsection{Evaluation Comparison}
\textbf{Results for face Verification on LFW, CALFW, CPLFW.} 
LFW \cite{huang2008labeled} dataset is the most widely used benchmark for unconstrained face verification on images. 
Recent algorithms\cite{deng2019arcface,wang2018support} gain nearly saturation performance on it.
Instead of only comparing performance on LFW, we also employed CALFW and CPLFW datasets, which involved higher pose and age variations for same identities from LFW, to evaluate the performance for face verification.
The experiment results were reported in Table \ref{verification_table}.
From the results, we could find that several algorithms gained better performance than human-individual. 
By comparing results from different algorithms, the results shown that although both Arcface and our proposed LATSE gain similar performance on LFW test, our method improved 1\% accuracy by average on CALFW and CPLFW benchmarks.
This showed the proposed method generalized better when more human pose and age variations were involved. 

\begin{table}[t]
\caption{Face verification results ($\%$) for different algorithms}
\label{verification_table}
\vskip 0.15in
\begin{center}
\begin{small}
%\begin{sc}
\begin{tabular}{lcccr}
\toprule
Method & LFW & CALFW & CPLFW \\
\midrule
\scriptsize{Human-Individual} & 97.27 & 82.32 & 81.21 \\
\scriptsize{Human-Fusion}     & $\mathbf{99.85}$ & 86.50 & 85.24 \\
\midrule
\scriptsize{CenterLoss \cite{wen2016discriminative}}  & 98.75 & 85.48 & 77.48 \\
\scriptsize{SphereFace \cite{liu2017sphereface}}      & 99.27 & 90.30 & 81.40 \\
\scriptsize{VGGFace2 \cite{cao2018vggface2}}          & 99.43 & 90.37 & 84.00 \\
\scriptsize{ArcFace \cite{deng2019arcface}}           & $\mathbf{99.82}$ & 95.45 & 92.08 \\
\scriptsize{Proposed LATSE}                       & $\mathbf{99.82}$ & $\mathbf{96.20}$ & $\mathbf{93.48}$ \\
\bottomrule
\end{tabular}
%\end{sc}
\end{small}
\end{center}
\vskip -0.1in
\end{table}

\textbf{Results for MegaFace test.} The MegaFace \cite{kemelmacher2016megaface} dataset includes 1,000,000 images of 690,000 different identities as the distractors in the gallery set and 100,000 photos of 530 unique celebrity from FaceScrub as the probe set.
There are two testing scenarios, one for face identification and the other is face verification. 
The testing process is conducted under two different training protocols (large or small training set, while training set is defined as large if there is more than 500,000 images in it).
Following the setting in ArcFace \cite{deng2019arcface}, we employed CASIA as training dataset for protocol of small, and use MS1M-RetinaV dataset as the training set for protocol large.
For MegaFace test data inference, we used the cleaned version from ArcFace to test the proposed method, which is noted as `r' in the Table \ref{megaface_table}.
Both the identification and verification accuracy results were reported to compare the performance for large scale face recognition.
Our method improved both accuracy under small and large training protocols. 
The results showed that the proposed architecture gain more generalization capability for large scale face recognition in an open set situation. 
We gained more than $99\%$ accuracy on MegaFace test without private dataset.

\begin{table}[ht]
\caption{Megaface results ($\%$) for different algorithm}
\label{megaface_table}
\vskip 0.15in
\begin{center}
\begin{small}
%\begin{sc}
\begin{tabular}{lcccr}
\toprule
Method          & Identification  & Verification \\
\midrule
\scriptsize{Softmax \cite{liu2017sphereface}} & 54.85 &  65.92     \\
\scriptsize{TripletLoss \cite{schroff2015facenet}} & 64.79 & 78.32 \\
\scriptsize{SphereFace \cite{liu2017sphereface}} & 72.73 & 85.56 \\
\scriptsize{SphereFace+\cite{liu2018learning}}  & 73.03 & -     \\
\scriptsize{CosFace \cite{wang2018cosface}}      & 77.11      & 89.88 \\
\scriptsize{ArcFace(CASIA) \cite{deng2019arcface}} & 77.50  & 92.34 \\
\scriptsize{ArcFace,r(CASIA) \cite{deng2019arcface}} & 91.75 & 93.69 \\
\midrule
%\scriptsize{LinearFace on CASIA,R}    & & \\
%\scriptsize{LinearFace+TS on CASIA,R}    & & \\
%\scriptsize{LinearFace+GAN on CASIA,R}    & & \\
\scriptsize{Proposed LATSE,r(CASIA)}    & $\mathbf{94.74}$ & $\mathbf{95.38}$ \\
\midrule
\midrule
\scriptsize{TripletLoss \cite{schroff2015facenet}} & 70.49 & 86.47 \\
\scriptsize{CosFace \cite{wang2018cosface}}        & 82.72 & 96.65 \\
\scriptsize{ArcFace,r \cite{deng2019arcface}}     & 98.35 & 98.48 \\
\scriptsize{SV-AM-Softmax,r \cite{wang2018support}}  & 98.82 & 99.03   \\
\midrule
%\scriptsize{LinearFace,R}    & & \\
%\scriptsize{LinearFace+TS,R}    & & \\
%\scriptsize{LinearFace+GAN,R}    & & \\
\scriptsize{Proposed LATSE,r}    & $\mathbf{99.14}$ & $\mathbf{99.19}$ \\
\bottomrule
\end{tabular}
%\end{sc}
\end{small}
\end{center}
\vskip -0.1in
\end{table}

\section{Concluding Remarks}
In this paper, we illustrate how the margin based loss methods work for face embedding learning in a perspective of probability. 
Two principles are proposed for new margin loss function designing.
At the same time, in view of comparison metric encounters the bottleneck to gain more generalization ability, we propose to regard the open set face recognition as a problem of information transmission.
Based on this intuition, we propose an auto-encoder architecture trained with a teacher student learning strategy, which increased the generalization ability for the face embedding. 
The extensive experimental results on several benchmarks show clear advantages of the proposed method.

\bibliography{icml_face}
\bibliographystyle{icml2020}

\end{document}